%% file: paper.tex
\documentclass[conference]{IEEEtran}
\IEEEoverridecommandlockouts

\usepackage{cite}
\usepackage{amsmath,amssymb,amsfonts}
\usepackage{algorithmic}
\usepackage{graphicx}
\usepackage{textcomp}
\usepackage{xcolor}

\usepackage{tikz}
\usepackage{booktabs}       
\usepackage{url}            
\usepackage[hidelinks]{hyperref} 
\usepackage{listings}       
\usepackage{multirow}       
\usepackage{balance}        
\usepackage[T1]{fontenc}    

\usepackage{pgfplots}
\pgfplotsset{compat=1.17} 
\usepackage{amsmath}

\definecolor{codegreen}{rgb}{0,0.6,0}
\definecolor{codegray}{rgb}{0.5,0.5,0.5}
\definecolor{codepurple}{rgb}{0.58,0,0.82}
\definecolor{backcolour}{rgb}{0.98,0.98,0.98}

\lstdefinestyle{mystyle}{
    backgroundcolor=\color{backcolour},
    commentstyle=\color{codegreen},
    keywordstyle=\color{magenta},
    numberstyle=\tiny\color{codegray},
    stringstyle=\color{codepurple},
    basicstyle=\ttfamily\footnotesize,
    breakatwhitespace=false,
    breaklines=true,
    captionpos=b,
    keepspaces=true,
    numbers=left,
    numbersep=5pt,
    showspaces=false,
    showstringspaces=false,
    showtabs=false,
    tabsize=2,
    frame=single, 
    rulecolor=\color{black!30} 
}
\def\BibTeX{{\rm B\kern-.05em{\sc i\kern-.025em b}\kern-.08em
    T\kern-.1667em\lower.7ex\hbox{E}\kern-.125emX}}

\usepackage{xspace}
\newcommand{\framework}{DURA-CPS\xspace}

\begin{document}

\title{DURA‑CPS: A Multi‑Role Orchestrator for Dependability Assurance in LLM‑Enabled Cyber‑Physical Systems

}
\author{%
  \IEEEauthorblockN{%
    Trisanth Srinivasan\textsuperscript{1*}, 
    Santosh Patapati\textsuperscript{1*}, 
    Himani Musku\textsuperscript{2}, 
    Idhant Gode\textsuperscript{3},\\
    Aditya Arora\textsuperscript{4}, 
    Samvit Bhattacharya\textsuperscript{1}, 
    Abubakr Nazriev\textsuperscript{5}, 
    Sanika Hirave\textsuperscript{6},\\
    Zaryab Kanjiani\textsuperscript{3}, 
    Srinjoy Ghose\textsuperscript{7}
  }
  \vspace{1ex}
  \IEEEauthorblockA{%
    \textsuperscript{1}Cyrion Labs\\
    \textsuperscript{2}Carnegie Mellon University\\
    \textsuperscript{3}Cornell University\\
    \textsuperscript{4}University of California San Diego\\
    \textsuperscript{5}University of Montana\\
    \textsuperscript{6}Oakland University\\
    \textsuperscript{7}University of Pennsylvania\\[1ex]
  }
}

\maketitle 

\begin{abstract}
Cyber-Physical Systems (CPS) increasingly depend on advanced AI techniques to operate in critical applications. However, traditional verification and validation methods often struggle to handle the unpredictable and dynamic nature of AI components. In this paper, we introduce \framework, a novel framework that employs multi-role orchestration to automate the iterative assurance process for AI-powered CPS. By assigning specialized roles (e.g., safety monitoring, security assessment, fault injection, and recovery planning) to dedicated agents within a simulated environment, \framework continuously evaluates and refines AI behavior against a range of dependability requirements. We demonstrate the framework through a case study involving an autonomous vehicle navigating an intersection with an AI-based planner. Our results show that \framework effectively detects vulnerabilities, manages performance impacts, and supports adaptive recovery strategies, thereby offering a structured and extensible solution for rigorous V\&V in safety- and security-critical systems.
\end{abstract}


\input{sections/01_introduction_cpsguard.tex}
\input{sections/02_related_work_cpsguard.tex}
\input{sections/03_framework_cpsguard.tex}
\input{sections/04_usecase_cpsguard.tex}
\input{sections/05_results_cpsguard.tex}
\input{sections/06_discussion_cpsguard.tex}

\input{sections/07_conclusion_cpsguard.tex}

\balance 
\bibliographystyle{IEEEtran}
\bibliography{IEEEabrv,references_cpsguard} 

\end{document}

%% file: sections/01_introduction_cpsguard.tex
\section{Introduction}
\label{sec:introduction}

Cyber-Physical Systems (CPS) integrate computational algorithms with physical processes. CPS is being used across various sectors such as transportation, energy, manufacturing, healthcare, and agriculture. Ensuring that these systems are safe, secure, reliable, and timely is critical as failures may result in serious consequences. Thus, Verification and Validation (V\&V) is essential to build trust that a CPS meets its requirements and behaves as intended. V\&V remains a complex and costly endeavor in most domains \cite{Klein2014}.

The challenges of V\&V grow when Artificial Intelligence (AI) is incorporated into CPS. This difficulty is compounded as increasingly complex AI techniques, such as Deep Neural Networks (DNNs) and Large Language Models (LLMs) are deployed in these systems. The behavior of AI is often sensitive to unexpected changes in the environment \cite{Lekadir2024}. This unpredictability, coupled with a vulnerability to adversarial attacks \cite{Goodfellow2015}, makes traditional V\&V methods, which rely on predictable behavior and exhaustive state exploration \cite{Clarke1999}, less effective for assessing AI-based components.

Addressing these issues requires V\&V frameworks designed specifically for AI-based CPS. Such frameworks should validate AI models in isolation and how they interact with the environment and other system components at runtime. We require structured approaches that can systematically:

\begin{itemize}
    \item Evaluate AI behavior in context against diverse dependability criteria
    \item Assess robustness to noise, faults, and security threats
    \item Facilitate the analysis of interactions between AI, the physical system, and its environment
    \item Support iterative refinement
    \item Provide traceable evidence suitable for building assurance cases and supporting certification \cite{Hawkins2011AssuranceCases}
\end{itemize}

To meet these needs, we present \framework, a novel framework centered around multi-role orchestration to iteratively assure the dependability of AI components within CPS. \framework assigns roles to various agents (AI models, algorithms, formal checkers, etc.) to work together within a simulated CPS environment. The agents continuously assess, challenge, and refine the performance of the primary AI component according to the defined dependability requirements.

The contributions of the proposed framework are as follows:

\begin{enumerate}
    \item \textbf{Multi-Role V\&V Architecture}: Defines specialized roles (e.g., 'Generator', 'SafetyMonitor', 'SecurityAssessor', 'PerformanceOracle', 'FaultInjector', 'RecoveryPlanner') tailored for comprehensive dependability assessment.
    \item \textbf{Iterative Assurance Loop}: Implements a controlled feedback loop where V\&V findings from monitoring/assessment roles inform subsequent actions, test generation, or adaptation planning.
    \item \textbf{Simulation Integration \& State Management:} Provides interfaces for connecting to standard CPS simulators and manages the shared state information needed for contextual V\&V.
    \item \textbf{Extensibility:} Designed for modularity, allowing users to define new roles or customize existing ones with different AI models, formal techniques, or procedural logic.
    \item \textbf{Dependability-Focused Metrics:} Collects metrics specifically related to safety violations, security vulnerabilities detected, performance adherence, and recovery effectiveness.
\end{enumerate}

By orchestrating these roles in a coordinated manner, \framework aims to provide a more rigorous approach to V\&V for AI in CPS than ad-hoc testing or isolated component analysis. 

%% file: sections/02_related_work_cpsguard.tex
\section{Related Work}
\label{sec:related_work}

The assurance of AI-based CPS draws upon research within several fields.

\subsection{Verification and Validation of CPS}

Traditional V\&V approaches for CPS are based on formal methods such as model checking and theorem proving \cite{Alur2015PrinciplesCPS}, simulation-based testing \cite{SimulationTestingCPS}, hardware-in-the-loop validation, runtime verification \cite{Leucker2009RVBook, Basin2018RVforCPS}, and fault or attack injection \cite{FaultInject_Ref}. These methods form the backbone of CPS assurance. However, they face challenges when applied to complex AI components because of issues with scalability, lack of test coverage, and difficulties in modeling the unpredictable behavior of AI \cite{Lekadir2024}. 

\subsection{Verification and Validation of AI/ML Systems}

Research on AI V\&V focuses on robustness testing against perturbations and adversarial examples \cite{Carlini2017Towards, Nicolae2018ART}, as well as the formal verification of network properties using techniques such as SMT solvers or abstract interpretation \cite{Katz2017Reluplex, Gehr2018AI2}. Efforts have also been made to improve explainability \cite{Ribeiro2016LIME, Lundberg2017SHAP} and assess fairness in machine learning systems \cite{Fairness_Ref}. Simulation-based test generation is also commonly used to supplement offline analyses \cite{Tuncali2018SimulationTestingAV, Abdessalem2018TestingAV}. However, these approaches tend to target specific individual properties. They do not capture the full complexity of runtime interactions in a dynamic CPS environment, such as chain reaction events \cite{uuk2024taxonomy}.

\subsection{Runtime Assurance and Monitoring}

Runtime Assurance (RA) techniques aim to ensure safety during operation by employment monitors and safety controllers \cite{Sha2001Simplex, Falcone2018RVHandbook}. For example, the Simplex architecture \cite{Sha2001Simplex} switches between a complex primary controller and a simpler, verified safety controller as needed. Runtime verification methods assess execution traces against formal specifications using temporal logic such as LTL, MTL, or STL \cite{Maler2004STL, Donze2013Breach}. In contrast to these approaches, \framework utilizes multiple coordinated roles in a closed-loop assurance process. This goes beyond just switching to backup controllers.

\subsection{Multi-Agent Systems (MAS) for Simulation and V\&V}

Multi-agent systems have been used to simulate complex systems and test control strategies in CPS \cite{MAS_Simulation_Survey}. Agent-based modeling has been used to mimic CPS behavior which focuses on autonomous driving scenarios. These approaches simply utilize agent interactions for simulation. On the other hand, \framework assigns specific V\&V roles that target dependability assurance rather than general-purpose simulation or emergent behavior analysis.

\subsection{LLM and AI Orchestration Frameworks}

Recent frameworks such as LangChain \cite{LangChain_Ref} and LlamaIndex \cite{LlamaIndex_Ref} have shown how to compose LLM calls with external tools and data to build applications. AutoGen \cite{AutoGen_Ref} further supports complex workflows involving multiple LLM agents. Although these orchestration frameworks demonstrate considerable power, they are primarily designed for application development and general AI collaboration. Their capabilities for formal requirement checking, CPS simulation, and structured assurance loops are limited compared to the dedicated design of \framework. This often forces researchers to develop custom assurance loops and V\&V steps from the ground up, which is highly time consuming.  Earlier work, such as the LLMOrchestrator \cite{llmorchestrator2024}, provided a basic generator-verifier structure that \framework significantly expands by incorporating multiple V\&V roles.

\subsection{Positioning \framework}

\framework builds on ideas from the aforementioned fields while establishing a new niche. It adopts the orchestration paradigm from AI frameworks, re-purposing it specifically for dependability V\&V. The frameworks employs a multi-role architecture inspired by multi-agent systems. It assigns roles that focus on safety monitoring, security testing, performance evaluation, fault injection, and recovery planning. Unlike approaches that rely solely on runtime monitoring and simple controller switching, \framework implements an iterative, closed-loop process that integrates tightly with CPS simulation environments as a robust solution for the complex challenges of V\&V in AI-based CPS.

%% file: sections/03_framework_cpsguard.tex
\section{The \framework Framework}
\label{sec:framework}

\framework is a Python-based framework designed to structure and automate the iterative V\&V process for AI components within simulated CPS environments. It employs a multi-role orchestration paradigm where specialized computational agents, termed Roles, collaborate to assess the behavior of the primary AI component Under Test (AUT) against dependability requirements.

\subsection{Core Architecture}
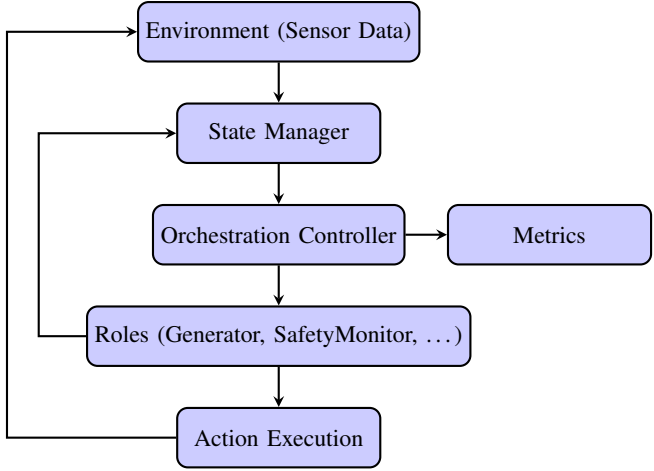
\begin{figure}[htbp]
\centering
\begin{tikzpicture}[
  node distance=1.5cm, 
  auto, 
  scale=0.9, 
  transform shape,
  block/.style={
    rectangle, 
    draw, 
    fill=blue!20, 
    thick, 
    text centered, 
    rounded corners, 
    minimum height=2.5em, 
    minimum width=3cm
  },
  arrow/.style={->,>=stealth, thick}
]

\node[block] (env) {Environment (Sensor Data)};
\node[block, below of=env] (state) {State Manager};
\node[block, below of=state] (orch) {Orchestration Controller};
\node[block, below of=orch] (roles) {Roles (Generator, SafetyMonitor, \dots)};
\node[block, below of=roles] (action) {Action Execution};

\node[block, right of=orch, xshift=2.5cm] (metrics) {Metrics};

\draw[arrow] (env) -- (state);
\draw[arrow] (state) -- (orch);
\draw[arrow] (orch) -- (roles);
\draw[arrow] (roles) -- (action);

\draw[arrow] (roles.west) -- ++(-0.7,0) |- (state.west);

\draw[arrow] (action.west) -- ++(-2.5,0) coordinate(temp3)
             -- (temp3 |- env.west) coordinate(temp4)
             -- (env.west);

\draw[arrow] (orch.east) -- (metrics.west);

\end{tikzpicture}
\caption{Overview of the \framework architecture. Sensor data from the CPS is collected by the Environment Interface and organized by the State Manager. The Orchestration Controller coordinates specialized Roles that generate and refine actions through a dedicated Action Execution module. These actions are fed back into the system to complete a closed-loop assurance process, while Metrics are concurrently tracked for continuous verification and validation}
\label{fig:architecture_cpsguard}
\end{figure}

The \framework architecture, illustrated in Fig. \ref{fig:architecture_cpsguard}, centers around an Orchestration Controller that manages the interaction between various Roles, an Environment Interface connecting to the CPS simulator, a State Manager maintaining shared state, and a Dependability Metrics tracker.

\subsection{Key Components}

\framework consists of five key components.

\subsubsection{Orchestration Controller}
This central component manages the overall execution flow. It initializes the roles, manages the iterative V\&V loop, sequences role execution based on dependencies or triggers, facilitates communication between roles (via the State Manager), and terminates the process based on predefined criteria (e.g., number of iterations, test completion, violation detected).

\subsubsection{Role}
A Role represents a specialized function within the V\&V process. It's an abstract base class defining a standard interface. Users implement or configure concrete Role subclasses. Roles interact indirectly through a State Manager. Key predefined (but extensible) roles include:
\begin{itemize}
    \item \textbf{Generator:} Represents the primary AI component Under Test (AUT) or a component generating inputs/scenarios for it. Takes current state/context, generates an action, plan, or output.
    
    \item \textbf{SafetyMonitor:} Checks the state, proposed actions, or predicted outcomes against safety rules or invariants. Can use rule-based logic, formal specifications (e.g., STL checks via integrated monitors like RTAMT \cite{Nickovic2016RTAMT}), or even another AI model trained for safety assessment. Returns a safety verdict (e.g., safe, unsafe, warning) and potentially quantitative scores.
    
    \item \textbf{SecurityAssessor:} Evaluates the system's security posture. Can analyze potential vulnerabilities based on the current state or AI output, or direct the FaultInjector. Might involve checking against known attack patterns or security policies.
    
    \item \textbf{PerformanceOracle:} Monitors performance metrics against requirements (e.g., response time, resource usage, control accuracy, task completion metrics).
    
    \item \textbf{FaultInjector:} Introduces faults or disturbances into the simulation based on directives (e.g., from the SecurityAssessor or predefined test plans). Can simulate sensor noise/failure, communication delays/loss, GPS spoofing, or adversarial perturbations to AI inputs.
    
    \item \textbf{RecoveryPlanner:} Activated upon detection of safety/security violations or critical failures. Proposes recovery actions or adaptations (e.g., switch to safe mode, replan trajectory, trigger alert). Can be rule-based or use planning algorithms/AI.
\end{itemize}

\subsubsection{Environment Interface}
Provides an abstraction layer for communicating with the external CPS simulator (e.g., CARLA, AirSim, Gazebo \cite{Dosovitskiy2017CARLA, Shah2017, Koenig2004}). It handles:
\begin{itemize}
    \item Sending commands/actions (from Generator or RecoveryPlanner) to the simulator.
    \item Receiving sensor data and state updates from the simulator.
    \item Translating between the simulator's data formats and the internal state representation.
    \item Potentially controlling simulation time steps or scenario loading.
\end{itemize}

\subsubsection{State Manager}
Maintains the shared state accessible by all roles. This includes:
\begin{itemize}
    \item Current state received from the Environment Interface (e.g., vehicle position, sensor readings).
    \item Outputs produced by roles in the current iteration (e.g., the Generator's proposed action, the SafetyMonitor's verdict, the FaultInjector's active fault).
    \item Historical state information if needed for temporal analysis.
\end{itemize}
Ensures consistent view of the system state for all roles within an iteration.

\subsubsection{DependabilityMetrics}
Collects and logs key metrics throughout the orchestration process, such as:
\begin{itemize}
    \item Number and type of safety/security violations detected.
    \item Performance metric values over time.
    \item Robustness scores from monitors (if applicable).
    \item Fault injection success/impact.
    \item Recovery action success rates.
    \item Processing time per role/iteration.
\end{itemize}
This data is crucial for post-hoc analysis and generating assurance reports.

\subsection{Iterative Orchestration Workflow}
The \framework workflow can be customized as needed. A typical execution cycle proceeds as follows:

\begin{enumerate}
    \item \textbf{Initialization:} Controller loads configuration, initializes roles, connects to the simulator via the Environment Interface, and gets initial state via the State Manager.
    \item \textbf{Iteration Start:} Controller triggers roles based on sequence or dependencies.
    \item \textbf{State Update:} The Environment Interface provides current world state to the State Manager.
    \item \textbf{Generation or Action Proposal:} The Generator (AUT) proposes an action based on current state.
    \item \textbf{Dependability Assessment:}
        \begin{itemize}
        \item SafetyMonitor evaluates proposed action/state against safety rules.
        \item SecurityAssessor evaluates security posture; may direct the FaultInjector.
        \item PerformanceOracle checks performance metrics.
        \item FaultInjector potentially introduces faults/attacks based on directives or test plan.
        \end{itemize}
    \item \textbf{V\&V Feedback Processing:} Controller gathers verdicts/outputs from assessment roles via the StateManager.
    \item \textbf{Decision and Adaptation:}
        \begin{itemize}
        \item If violations detected: Controller may halt, log details, or activate the RecoveryPlanner. The RecoveryPlanner proposes alternative action.
        \item If no violations: Controller approves Generator action (or refined action).
        \end{itemize}
    \item \textbf{Action Execution:} Controller sends the final approved or recovery action to the simulator via the EnvironmentInterface.
    \item \textbf{Metrics Logging:} Relevant data is logged through DependabilityMetrics for the iteration.
    \item \textbf{Loop/Terminate:} Controller checks termination conditions (e.g., time limit, scenario end, critical failure). If not met, proceeds to the next iteration (Step 2).
\end{enumerate}
This loop allows for continuous evaluation and adaptation, forming an iterative assurance process within the simulation. The configuration of roles, their interaction logic (dependencies and triggers), and the connection to the simulation environment define the specific V\&V experiment conducted using the framework.

\subsection{Extensibility}
\framework is designed for extensibility. Users can:
\begin{itemize}
    \item Implement new Role subclasses using custom Python code, integrating different AI models (LLMs, DNNs), formal verification tools (via wrappers), or standard algorithms.
    \item Define complex interaction protocols and triggering conditions between roles.
    \item Develop new EnvironmentInterface subclasses to support different simulators or hardware-in-the-loop setups.
    \item Customize the DependabilityMetrics collection.
\end{itemize}
This allows tailoring the framework to specific CPS domains, AI components, and V\&V requirements.

%% file: sections/04_usecase_cpsguard.tex
\section{Use Case: Autonomous Vehicles}
\label{sec:usecase}

\begin{figure}
\centering
\includegraphics[width=0.95\columnwidth]{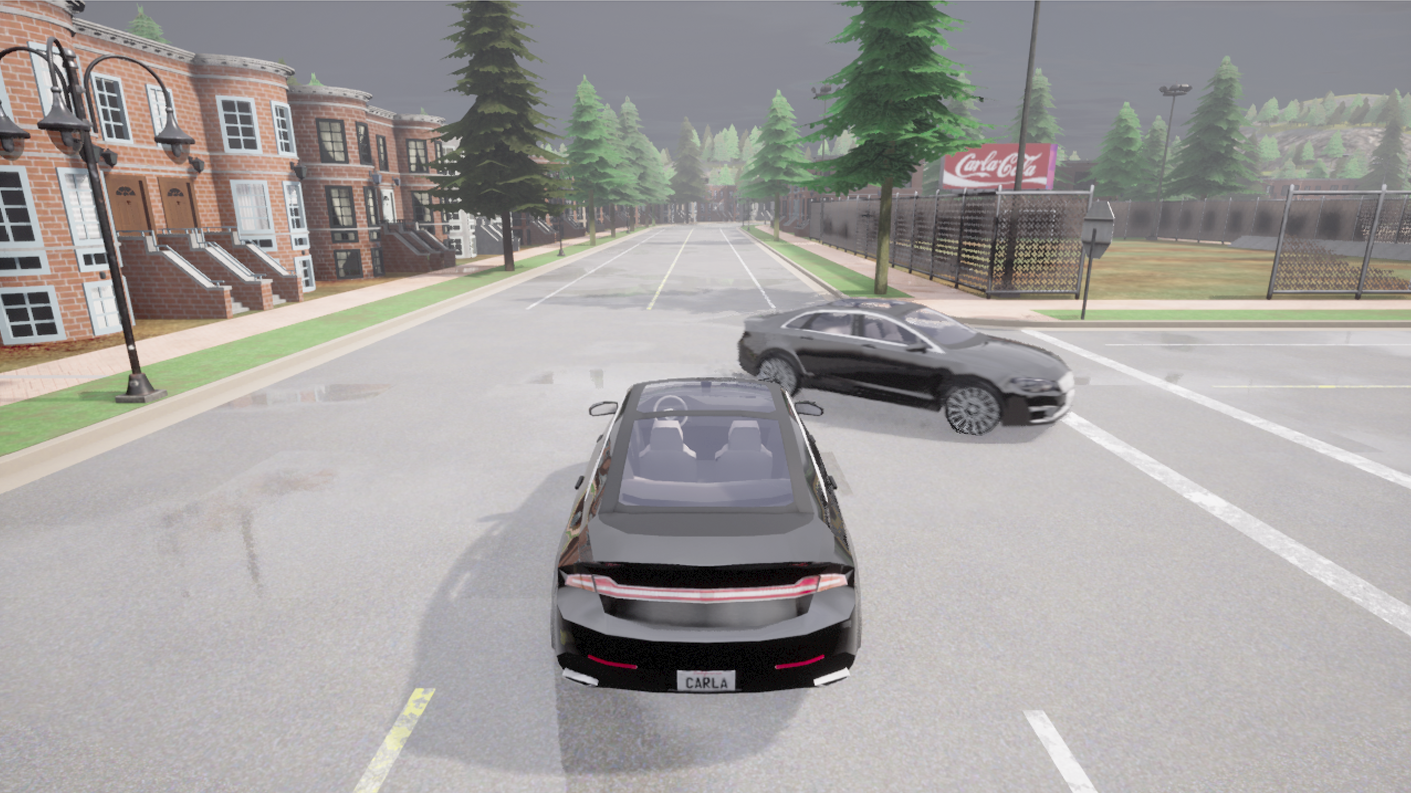}
\caption{A sample CARLA 3D scene. A third-person view was used as input for our Llama 3.2 11B model alongside sensor readings. \cite{grattafiori2024llama3, Dosovitskiy2017CARLA, li2023dialogue}}
\label{fig:SampleCARLA3DScene}
\end{figure}

To demonstrate the capabilities of \framework, we apply it to a challenging V\&V scenario: ensuring the safe and secure navigation of an unsignalized urban intersection by an autonomous vehicle (AV) equipped with an AI-based planning module.

\subsection{Scenario Description}

\begin{figure*}
\centering
\includegraphics[width=2\columnwidth]{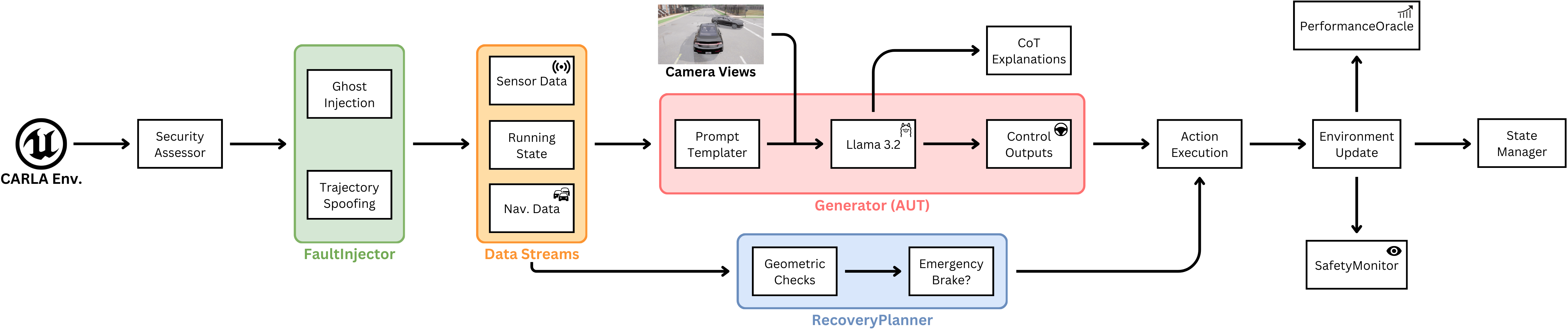}
\caption{High-level orchestration of AI-based generator and other roles within \framework. The CARLA environment supplies navigation (map, traffic, waypoints) and sensor data. The SecurityAssessor can inject faults through the FaultInjector. These data streams, alongside the running state, feed into a prompt templater to generate a textual representation for the Llama 3.2 11B model. Camera views are passed directly to the LLM. Llama 3.2 generates both control outputs and corresponding explanations. The Action Execution module then applies these outputs to the environment. In parallel, the RecoveryPlanner employs geometric checks and determines whether or not to employ the emergency brake, which overrides all other actions. The running state is updated via the StateManager to include past actions and associated CoT explanations. The PerformanceOracle and SafetyMonitor track for performance.}
\label{fig:UseCase-Architecture}
\end{figure*}

The AV must navigate a four-way intersection potentially shared with other vehicles (simulated background traffic) and pedestrians. The primary AI component Under Test (AUT) is an LLM-based tactical planner. Given sensor inputs (object lists, positions, velocities from simulated perception) and a high-level goal (e.g., "proceed straight"), the LLM planner generates maneuver decisions (e.g., "wait", "accelerate", "yield", "proceed cautiously"). Critical dependability requirements include:
\begin{itemize}
    \item \textbf{Safety:} Avoid collisions with other vehicles and pedestrians. Maintain safe following distances. Obey traffic rules implicitly (e.g., right-of-way, though not explicitly programmed).
    \item \textbf{Security:} Resilience against spoofed sensor data (e.g., ghost obstacles, false trajectories) intended to cause hazardous behavior or gridlock.
    \item \textbf{Performance:} Navigate the intersection without undue delay (avoiding excessive conservatism) or comfort violations (jerk/acceleration).
\end{itemize}

\subsubsection{Rationale for an LLM-Based Planner}
Existing AV planning stacks use domain-specific rule sets. While this is effective, it limits our ability to stress-test \framework across diverse AI use cases. We therefore deliberately use an LLM-based planner, which is relatively weaker in this area, to 1) show the framework can wrap any black-box decision module, 2) surface failure modes that traditional frameworks would not find, and 3) to demonstrate how DURA-CPS's different roles interact when the AI's internal logic is opaque.

\subsection{\framework Configuration}

We configured \framework with the following roles instantiated for this scenario (Figure \ref{fig:UseCase-Architecture}):

\begin{itemize}
    \item \textbf{Generator (AUT):} An LLM (fine-tuned Llama 3.2 11B variant \cite{grattafiori2024llama3}) prompted with current perceived world state and goal, outputting a tactical maneuver decision. The LLM is provided few-shot examples and a Chain-of-Thought (CoT) prompt. Table \ref{tab:sensorinputs} details the sensor inputs the LLM receives.
    \item \textbf{SafetyMonitor:} Implemented using geometric checks and simplified traffic rules. It verifies if the proposed maneuver maintains a minimum safety distance from all perceived dynamic objects based on predicted trajectories. Flags "unsafe" if violations are predicted.
    \item \textbf{SecurityAssessor:} Monitors incoming sensor data patterns. For this use case, it directs the FaultInjector to periodically introduce specific attacks.
    \item \textbf{FaultInjector:} Simulates two attack types based on SecurityAssessor triggers:
        \begin{enumerate}
            \item \textit{Ghost Obstacle Injection:} Adds a non-existent dynamic obstacle into the perceived state provided to the Generator.
            \item \textit{Trajectory Spoofing:} Modifies the predicted velocity or path of a real detected vehicle to appear more hazardous than it is.
        \end{enumerate}
    \item \textbf{PerformanceOracle:} Tracks intersection clearance time and maximum longitudinal/lateral acceleration/jerk. Flags "performance\_fail" if thresholds are exceeded.
    \item \textbf{RecoveryPlanner:} A simple rule-based agent. Using the same geometric checks as the SafetyMonitor, it flags checks for unsafe conditions. If unsafe conditions are detected, it overrides the Generator's decision with "emergency\_brake".
\end{itemize}

\begin{table*}[ht]
\centering
\caption{CARLA Sensor Inputs for Use Case Architecture}
\begin{tabular}{p{0.3\textwidth}p{0.65\textwidth}}
\toprule
\textbf{Sensor Input} & \textbf{Description} \\
\midrule
LiDAR-based Obstacle Summary & Textual summary of obstacles extracted from the LiDAR. Instead of using raw 3D data, the CarlaInterface aggregates nearby objects (vehicles, pedestrians, static obstacles) with positions \& dimensions. \\
Radar Summary & A text summary of radar detections that includes each object’s range and relative radial velocity. \\
Front RGB Camera & An RGB image captured from the front-facing camera passed directly to the LLM. \\
Third-Person View Camera & An RGB image providing a broader, third-person perspective of the intersection. This image delivers contextual clues about background traffic and environmental layout. \\
IMU Summary & A text-based summary of inertial measurements that includes linear acceleration, angular velocity, and heading. This information succinctly describes the vehicle’s motion dynamics. \\
Vehicle Speed & A numerical value representing the current speed of the vehicle, extracted from vehicle odometry. \\
HD Map \& Waypoint Data & A structured list of upcoming way points or lane center coordinates derived from a high-definition map. This input supports high-level route planning and navigation. \\
Traffic Controls Status & A concise textual report detailing the state of nearby traffic signals and the presence of key road signs. \\
\bottomrule
\end{tabular}
\label{tab:sensorinputs}
\end{table*}

\subsubsection{Integration}
The CARLA simulator \cite{Dosovitskiy2017CARLA}  was used via a custom \framework CarlaInterface (Figure \ref{fig:SampleCARLA3DScene}). The StateManager tracked perceived object lists (from CARLA's sensors), proposed actions, and associated CoT explanations.

\subsubsection{Orchestration Logic}
The controller executed roles sequentially within each simulated time step (100ms), where processing is aligned to 100 ms of simulated time, in the following order: Environment Update, Generator, SafetyMonitor, SecurityAssessor, FaultInjector (conditional), PerformanceOracle, Decision (Controller activates RecoveryPlanner if unsafe), Action Execution (Figure \ref{fig:architecture_cpsguard}).

\subsection{Test Scenarios}
We designed simulation scenarios with varying complexity and injected faults/attacks:
\begin{enumerate}
    \item \textbf{Nominal:} Light traffic, clear right-of-way.
    \item \textbf{Congested:} Moderate traffic density, requiring careful yielding and gap selection.
    \item \textbf{Conflicting Traffic:} Vehicles approaching simultaneously from multiple directions, testing navigation logic.
    \item \textbf{Ghost Obstacle Attack:} Nominal scenario + FaultInjector adds a ghost obstacle near the intersection entry.
    \item \textbf{Trajectory Spoofing Attack:} Congested scenario + FaultInjector spoofs the trajectory of an oncoming car to seem aggressive.
    \item \textbf{Pedestrian Crossing:} Scenario with a simulated pedestrian crossing the AV's intended path.
\end{enumerate}
Each scenario was run 15 times with variations in traffic patterns and timing.

\subsection{Expected Outcomes and Metrics}
We used \framework to assess:
\begin{itemize}
    \item \textbf{Safety Violations:} Frequency of the SafetyMonitor flagging "unsafe" maneuvers (indicating potential collisions based on geometry/rules). Actual collisions logged by CARLA serve as ground truth confirmation.
    \item \textbf{Security Resilience:} How the Generator (LLM) reacts to injected faults. Does it behave erratically, freeze (gridlock), or follow unsafe commands induced by fake data? Frequency of SafetyMonitor activations during attacks.
    \item \textbf{Performance Degradation:} Increase in intersection clearance time or comfort violations (jerk/acceleration) under congestion or attacks.
    \item \textbf{Recovery Effectiveness:} Success rate of the RecoveryPlanner (emergency brake) in preventing actual collisions when activated by the SafetyMonitor.
\end{itemize}
Metrics were collected by the DependabilityMetrics component.

%% file: sections/05_results_cpsguard.tex

\pgfplotsset{compat=1.17}

\section{Results and Analysis}
\label{sec:results}

This section presents the results from executing the autonomous intersection navigation use case with \framework across the defined scenarios (15 runs per scenario, total 90 runs).

\subsection{Safety Assessment}
The SafetyMonitor actively checked the LLM Generator's proposed maneuvers. Table \ref{tab:safety_violations} summarizes the percentage of runs where the SafetyMonitor flagged at least one "unsafe" prediction and the rate of actual collisions observed in CARLA.

\begin{table}[htbp]
\centering
\caption{Safety Monitor Activations and Collision Rates}
\label{tab:safety_violations}
\resizebox{\linewidth}{!}{
\begin{tabular}{@{}lccc@{}}
\toprule
Scenario Type           & Monitor Flags "Unsafe" (\%) & Collision Rate (\%) \\ \midrule
Nominal                 & 6.7\%  (1/15)  & 0.0\%  (0/15) \\
Congested               & 20.0\% (3/15)  & 6.7\%  (1/15) \\
Conflicting Traffic     & 33.3\% (5/15)  & 13.3\% (2/15) \\
Ghost Obstacle Attack   & 86.7\% (13/15) & 6.7\%  (1/15) \\ 
Trajectory Spoof Attack & 60.0\% (9/15)  & 20.0\% (3/15) \\ 
Pedestrian Crossing     & 26.7\% (4/15)  & 6.7\%  (1/15) \\ \midrule
\textbf{Overall Avg.}    & \textbf{38.9\%} & \textbf{8.9\%} \\ \bottomrule
\end{tabular}%
}
\vspace{-2mm}
\end{table}

Observations:
\begin{itemize}
    \item The LLM planner showed reasonable safety in nominal cases but struggled increasingly with complexity (Congested, Conflicting Traffic). It exposed weaknesses that validated the need for our multi-role loop.
    \item Security attacks significantly triggered the SafetyMonitor. The Ghost Obstacle attack frequently caused the LLM to propose sudden braking or swerving deemed unsafe by the monitor. Trajectory Spoofing often led the LLM to yield unnecessarily or hesitate, sometimes causing conflicts later flagged by the monitor.
    \item Actual collisions were lower than monitor flags, primarily because the RecoveryPlanner (emergency brake) often intervened successfully when triggered by the SafetyMonitor. The cases where collisions still occurred despite monitor flags often involved very short time-to-collision where braking was insufficient or complex multi-vehicle interactions.
\end{itemize}
\framework effectively identified scenarios where the LLM planner proposed potentially unsafe actions.

\subsection{Security Assessment and Resilience}
The FaultInjector, directed by the SecurityAssessor, successfully introduced faults. Analysis focused on the LLM's reaction:
\begin{itemize}
    \item \textbf{Ghost Obstacle:} The LLM consistently reacted to the ghost obstacle despite the visual input contradicting sensor input. It would often propose immediate braking or significant deceleration, treating it as real. This frequently led to performance issues (sudden stops, increased clearance time) and was often flagged by the SafetyMonitor if the braking was excessively abrupt. 
    
    \item \textbf{Trajectory Spoofing:} The LLM was sensitive to the spoofed aggressive trajectories, typically choosing to yield or wait much longer than necessary, significantly impacting performance. In 3 runs (20\%), this excessive caution led to situations where the AV became 'stuck', unable to find a perceived safe gap, resulting in a gridlock scenario broken only by simulation timeout.
\end{itemize}
\framework's roles allowed systematic injection and observation of the AI's vulnerability to sensor data manipulation.

\subsection{Performance Impact}
\begin{figure}[htbp]
\centering
\begin{tikzpicture}[scale=0.935, transform shape]
\begin{axis}[
    ybar,
    bar width=12pt,
    enlarge x limits=0.15,
    width=1.1\columnwidth,
    height=7cm,
    ylabel={Average Clearance Time (s)},
    symbolic x coords={
       Nominal,
       Congested,
       {Conflicting Traffic},
       {Ghost Obstacle Attack},
       {Trajectory Spoof Attack},
       {Pedestrian Crossing}
    },
    xtick=data,
    xticklabel style={rotate=30, anchor=east},
    axis on top,
    ymin=0, ymax=25, 
    nodes near coords,
    nodes near coords align={vertical},
    every node near coord/.append style={
      yshift=5pt,
      anchor=south,
      font=\small
    },
    error bars/.cd={ y dir=both, y explicit },
    every error bar/.append style={draw=red, line width=1pt, opacity=0.3},
    every error mark/.append style={rotate=90, yshift=2pt, opacity=1},
]
\addplot+[error bars/.cd, y dir=both, y explicit] 
coordinates {
    (Nominal,5.4)                +- (0,0.5)
    (Congested,9.9)              +- (0,1.0)
    (Conflicting Traffic,12.8)   +- (0,1.2)
    (Ghost Obstacle Attack,16.0) +- (0,4.8)
    (Trajectory Spoof Attack,18.4) +- (0,5.3)
    (Pedestrian Crossing,11.7)   +- (0,0.9)
};
\end{axis}
\end{tikzpicture}
\caption{Average Intersection Clearance Time Across Scenarios. Error bars indicate the standard deviation over 15 simulation runs.}
\label{fig:clearance_time}
\end{figure}
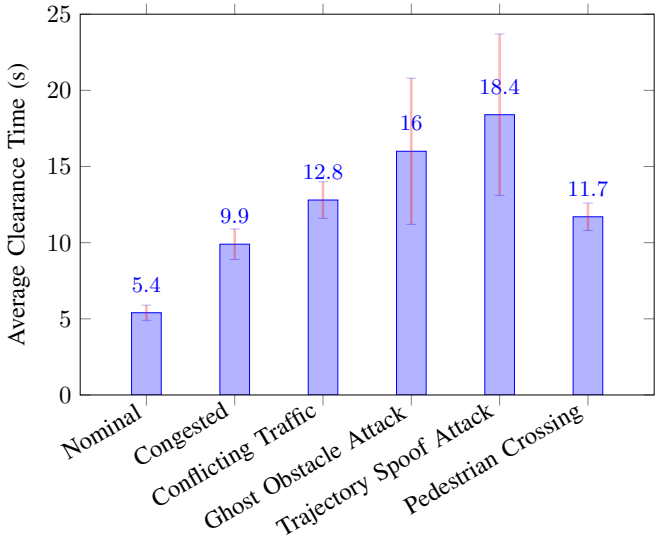

The PerformanceOracle tracked intersection clearance time and comfort metrics. 
Figure \ref{fig:clearance_time} shows average clearance time.

As expected, congestion and conflicting traffic increased clearance time. Security attacks had a major impact: Ghost Obstacles caused sharp braking, sometimes increasing time due to recovery, while Trajectory Spoofing significantly increased waiting times due to the LLM's overly cautious reaction to the fake aggressive behavior. Comfort violations (high jerk/acceleration) were most frequent during recovery braking and reactions to ghost obstacles.

\subsection{Recovery Effectiveness}
The simple RecoveryPlanner (emergency brake) was triggered whenever the SafetyMonitor flagged "unsafe".
\begin{itemize}
    \item It successfully prevented a collision in cases where it was activated and a collision would likely have occurred otherwise (based on manual inspection of near-miss scenarios).
    \item Failures typically occurred when the unsafe situation developed too rapidly for braking alone to suffice or involved complex side-impact scenarios.
\end{itemize}
This highlights the importance of the monitor-recovery loop but also suggests the need for more sophisticated recovery strategies than simple braking in future work.

\subsection{\framework Contribution Analysis}
This use case shows how \framework facilitates 
V\&V:
\begin{itemize}
    \item The multi-role setup allowed assessment of safety, security, and performance aspects of the LLM planner.
    \item The FaultInjector and SecurityAssessor enabled systematic testing against specific attack vectors.
    \item The closed-loop orchestration revealed interactions, such as security attacks leading to safety monitor triggers, or recovery actions impacting performance.
    \item DependabilityMetrics provided quantitative data summarizing complex behaviors across runs and scenarios.
\end{itemize}
The framework provided a systematic way to challenge the AI component and evaluate its dependability. 

%% file: sections/06_discussion_cpsguard.tex
\section{Discussion}
\label{sec:discussion}

The results from the autonomous intersection navigation use case provide insights into the challenges of assuring AI dependability in complex CPS and demonstrate the utility of the \framework framework's multi-role orchestration approach.

\subsection{Implications for AI/LLM Assurance in CPS}
The experiment shows the brittleness of even sophisticated AI models like LLMs when faced with complex 
scenarios and threats. The LLM planner showed degraded safety and performance under congestion, conflicting goals, and especially under simulated attacks. Its sensitivity to spoofed data (ghost obstacles and manipulated trajectories) is a significant concern, showing that relying solely on the AI's perceived world model without robust validation mechanisms is high-risk.

Security attacks directly impacted safety (by inducing unsafe reactions or hesitation leading to secondary conflicts) and performance (by causing excessive caution or gridlock). This demonstrates the need for V\&V frameworks like \framework that can assess these attributes concurrently and analyze their interactions, rather than treating them in isolation. The partial success of the simple recovery mechanism shows the importance of runtime assurance loops but also shows the need for more advanced monitoring and recovery strategies tailored to specific failure modes.

\subsection{Effectiveness and Role of \framework}
\framework proved effective in structuring this complex V\&V task. Its key strengths observed in the use case include:
\begin{itemize}
    \item \textbf{Structured Assessment:} Assigning specific dependability concerns (safety, security, performance) to distinct roles provided clarity and allowed for modular implementation of checks and attacks.
    \item \textbf{Systematic Fault/Attack Injection:} The FaultInjector role enabled controlled introduction of security threats, allowing systematic evaluation of the AI's resilience.
    \item \textbf{Closed-Loop Analysis:} The framework captured the dynamic feedback loop where AI actions influence the environment, which influences subsequent AI inputs and V\&V assessments, including 
    recovery actions.
    \item \textbf{Extensibility Potential:} While simple monitors and recovery were used here, the role-based structure readily accommodates more sophisticated implementations (e.g., STL-based monitors, AI-based security assessors).
\end{itemize}
\framework acts as an "in-the-loop V\&V orchestrator," allowing engineers to configure a virtual team of specialized agents that continuously probe and assess the AI AUT within its simulated operational context. This approach facilitates identifying weaknesses, understanding interactions between dependability facets, and evaluating the effectiveness of assurance mechanisms like monitors and recovery planners.

\subsection{Limitations and Future Work}
This work has several limitations that suggest avenues for future research:
\begin{itemize}
    \item \textbf{Specification Effort:} Defining logic for each role (especially monitors and assessors) requires great effort and domain expertise. Developing libraries of reusable role implementations would improve usability. 
    \item \textbf{Sim-to-Real Gap:} As with all simulation-based V\&V, transferring findings to the real world requires caution. Validating \framework and role implementations on physical platforms or high-fidelity digital twins is crucial.
    \item \textbf{Scalability:} Orchestrating many complex roles, especially those involving computationally intensive analysis or multiple AI inferences per time step, could become a bottleneck. 
    In the simulated environment, we executed the AI agents at each simulated interval rather than in real-time. Such an approach would not be possible in direct real-world testing.
    In such cases, performance optimization and distributed execution may be needed.
    \item \textbf{LLM Specifics:} While an LLM was used as the AUT, deeper analysis of LLM-specific failure modes (e.g., hallucination, prompt sensitivity) within the CPS context could be integrated into specific roles.
\end{itemize}

Future directions focus on:
\begin{enumerate}
    \item Developing a richer library of predefined V\&V roles incorporating diverse techniques (STL monitoring, simplified formal methods, ML-based anomaly detection). One approach may be to automatically generate V\&V roles using LLMs given the problem and constraints.
    \item Integrating \framework with hardware-in-the-loop (HIL) setups to evaluate its effectiveness.
    \item Applying \framework to other domains, including industrial robotics and dependable agricultural automation (e.g., safe human-robot collaboration in smart farming).
    \item Investigating optimizations and removing bottlenecks that limit \framework's effectiveness in out-of-sim tests.
    \item Developing specialized assessment metrics tailored to LLM-specific failure modes, such as hallucination, and integrating these into the \framework framework to improve LLM-based component assurance.
\end{enumerate}

%% file: sections/07_conclusion_cpsguard.tex
\section{Conclusion}
\label{sec:conclusion}

This paper introduces \framework, a framework leveraging multi-role orchestration to structure and automate the iterative assurance process for AI-based CPS. By assigning specialized V\&V functions to roles within a simulated CPS environment, \framework systematically evaluates AI behavior against a range of dependability requirements. Its iterative and closed-loop approach allows analysis of complex interactions and continuous evaluation of runtime assurance mechanisms.

We demonstrated the utility of \framework in a case study on autonomous vehicle intersection navigation with an LLM-based planner. The case study shows how \framework can effectively inject faults, detect safety and security vulnerabilities, assess performance impacts, and evaluate recovery actions. \framework provides quantitative metrics for assurance.

In conclusion, \framework offers a practical method for the V\&V of AI in CPS where safety and security are critical. Further work should focus on the identified challenges like specification effort and bridging the simulation-to-reality gap.